\documentclass[runningheads]{llncs}
\usepackage{graphicx}
\usepackage[width=122mm,left=12mm,paperwidth=146mm,height=193mm,top=12mm,paperheight=217mm]{geometry}
\usepackage{amsmath,amssymb} 
\usepackage{color}
\usepackage{sidecap}
\usepackage{cite}

\begin{document}

\title{Tracking Emerges by Colorizing Videos}

\titlerunning{Tracking Emerges by Colorizing Videos}

\authorrunning{Vondrick, Shrivastava, Fathi, Guadarrama, Murphy}

\author{Carl Vondrick, Abhinav Shrivastava, Alireza Fathi,\\Sergio Guadarrama, Kevin Murphy}
\institute{Google Research}

\maketitle              
\begin{abstract}
We use large amounts of unlabeled video to learn models for visual tracking without manual human supervision. We leverage the natural temporal coherency of color to create a model that learns to colorize  gray-scale videos by copying colors from a reference frame. Quantitative and qualitative experiments suggest that this task causes the model to automatically learn to track visual regions. Although the model is trained without any ground-truth labels, our method learns to track well enough to outperform the latest methods based on optical flow. Moreover, our results suggest that failures to track are correlated with failures to colorize, indicating that advancing video colorization may further improve self-supervised visual tracking. 
\keywords{colorization, self-supervised learning, tracking, video} 
\end{abstract}

\newcommand{\eat}[1]{} 
\newcommand{\KM}[1]{{\color{blue}{KM: #1}}}

\section{Introduction}

Visual tracking is integral for video analysis tasks across recognition, geometry, and interaction. However, collecting the large-scale tracking datasets necessary for high performance often requires extensive effort that is impractical and expensive. We believe a promising approach is to learn to track without human supervision by instead leveraging large amounts of raw, unlabeled video.



\begin{figure}[t]
    \centering
    \includegraphics[width=0.9\linewidth]{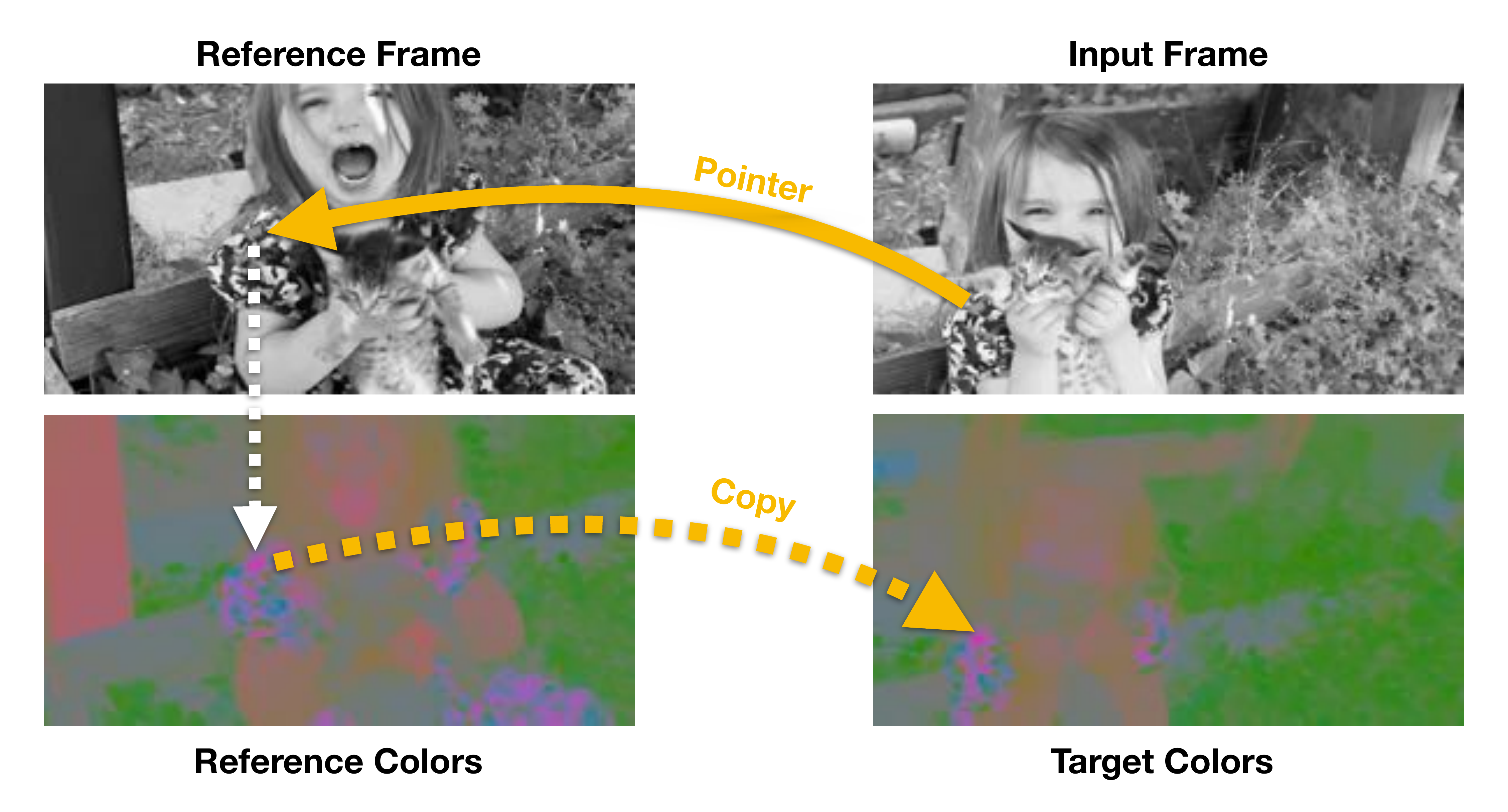}
    \caption{\textbf{Self-supervised Tracking:} We capitalize on large amounts of unlabeled video to learn a self-supervised model for tracking. The model learns to predict the target colors for a gray-scale input frame by pointing to a colorful reference frame, and copying the color channels. Although we train without ground-truth labels, experiments and visualizations suggest that tracking emerges automatically in this model.}
    \label{fig:teaser}
\end{figure}

We propose video colorization as a self-supervised learning problem for  visual tracking. 
However, instead of trying to predict the color directly from the gray-scale frame, we constrain the colorization model to solve this task by learning to copy colors from a reference frame. Although this may appear to be a roundabout way to colorize video, it requires the model to learn to internally point to the right region in order to copy the right colors.  Once the model is trained, the learned ``pointing'' mechanism acts as a tracker across time. Figure  \ref{fig:teaser} illustrates our problem setup.

Experiments and visualizations suggest that, although the network is trained without ground-truth labels, a mechanism for tracking automatically emerges. After training on unlabeled video collected from the web \cite{kay2017kinetics}, the model is able to track any segmented region
specified in the first frame of a video \cite{pont20182018}. It can also  track human pose  given keypoints annotated in an initial frame \cite{Jhuang:ICCV:2013}. While there is still no substitute for cleanly labeled supervised data, our colorization model learns to track video segments and human pose well enough to outperform the latest methods based on optical flow. Breaking down performance by motion type suggests that the colorization model is more robust than optical flow for many natural complexities, such as dynamic backgrounds, fast motion, and occlusions.  

A key feature of our model is that we do not require any labeled data during training. Our hypothesis, which our experiments support, is that learning to colorize video will cause a tracker to internally emerge, which we can directly apply to downstream tracking tasks without additional training nor fine-tuning. Moreover, we found that the failures from our tracker are often correlated with failures to colorize the video, which suggests that further improving our video colorization model can advance progress in self-supervised tracking.

The main contribution of this paper is to show that learning to colorize video causes tracking to emerge. The remainder of this paper describes this contribution in detail. In section 2, we first review related work in self-supervised learning and tracking. In section 3, we present our approach to use video colorization as a supervisory signal for learning to track. By equipping the model with a pointing mechanism into a reference frame, we learn an explicit representation that we can use for new tracking tasks without further training. In section 4, we show several experiments to analyze our method. Since annotating videos is expensive and tracking has many applications in robotics and graphics, we believe learning to track with self-supervision can have a large impact. 

\section{Related Work}


\textbf{Self-supervised Learning:} Our paper builds upon a growing body of work to train visual models without human supervision. A common approach is to leverage the natural context in images and video in order to learn deep visual representations \cite{doersch2015unsupervised,owens2016ambient,jayaraman2015learning,doersch2017multi,wang2017transitive,zhang2017split,larsson2017colorization,pathak2016context,wang2015unsupervised,vondrick2016generating,noroozi2016unsupervised,pathak2017learning,isola2017image}, which can be used as a feature space for training classifiers for down-stream tasks, such as object detection. Other approaches include interaction with an environment to learn visual features \cite{pinto2016curious,agrawal2016learning,wu2016physics}, which is useful for applications in robotics. A related but different line of work explores how to learn geometric properties or cycle consistencies with self-supervision, for example for motion capture or correspondence \cite{tung2017self,zhou2017unsupervised,zhou2016learning,ilg2017flownet,zhou2016view}.  We also develop a self-supervised model, but our approach focuses on visual tracking in video for segmentation and human pose. Moreover, our method is trained directly on natural data without the use of computer generated graphics \cite{zhou2016learning,ilg2017flownet}. 

\textbf{Colorization:} The task of colorizing gray-scale images has been the subject of significant study in the computer vision community \cite{welsh2002transferring,gupta2012image,liu2008intrinsic,chia2011semantic,deshpande2015learning,zhang2016colorful,larsson2016learning,guadarrama2017pixcolor,iizuka2016let,ironi2005colorization}, which inspired this paper. Besides the core problem of colorizing images, colorization has been shown to be a useful side task to learn representations for images without supervision \cite{zhang2017split,larsson2017colorization}. The task of colorization also been explored in the video domain \cite{yatziv2006fast,heu2009image} where methods can explicitly incorporate optical flow to provide temporal coherency or learn to propagate color \cite{liu2018switchable}. In this paper, we do not enforce temporal coherency; we instead leverage it to use video colorization as a proxy task for learning to track. 


\textbf{Video Segmentation:} One task that we use our tracker for is video segmentation where the task is to densely label object instances in the video. Methods for video segmentation are varied, but can generally be classified into whether they start with an object of interest \cite{badrinarayanan2010label,ramakanth2014seamseg,vijayanarasimhan2012active,perazzi2015fully} or not \cite{grundmann2010efficient,xu2012evaluation,brox2010object,fragkiadaki2012video}. The task is challenging, and state-of-the-art approaches typically use a large amount of supervision to achieve the best results \cite{yang2018efficient,caelles2017one,perazzi2017learning}, such as from ImageNet \cite{deng2009imagenet}, MS-COCO \cite{lin2014microsoft}, and DAVIS \cite{pont20182018}. We instead learn to track from just unlabeled video.

\textbf{Tracking without Labels:}
We build off pioneering work for learning to segment videos without labels \cite{faktor2014video,marki2016bilateral,khoreva2017lucid}. 
However, rather than designing a tracking objective function by hand, we show that there is a  self-supervised learning problem that causes the model to automatically learn tracking on its own. Consequently, our model is a \emph{generic tracking method} that is applicable to multiple video analysis problems and not limited to just video segmentation. The same trained model can track segments, track key points, colorize video, and transfer any other annotation from the first frame to the rest of the video, without any fine-tuning or re-training. To highlight that our tracker is generic, we show results for three materially different tracking tasks (colorization, video segmentation, keypoint tracking). Moreover, our approach is fast, tracks multiple objects, and does not require training on the testing frames, making our approach fairly practical for large-scale video analysis tasks.

\textbf{Note on Terminology:}
There is some disagreement in the tracking literature on terms, and we wish to clarify our nomenclature. In tracking, there are two common tasks. In task A, we are given the labels for the first frame. In task B, we are not given a labeled initial frame. The literature typically calls task A ``semi-supervised'' and task B ``unsupervised'' referring to whether the initial frame is labeled or not. The confusing terminology is that, in both cases, you are allowed to train with supervised data, even for the unsupervised task. In this paper, our goal is to \emph{learn only from unlabeled video}. At test time, we tackle task A, which specifies the region of interest to track. However, we call our method unsupervised because we do not learn with any labeled data.



\section{Self-supervised Tracking}

\begin{figure}[t]
    \centering
    \includegraphics[width=\linewidth]{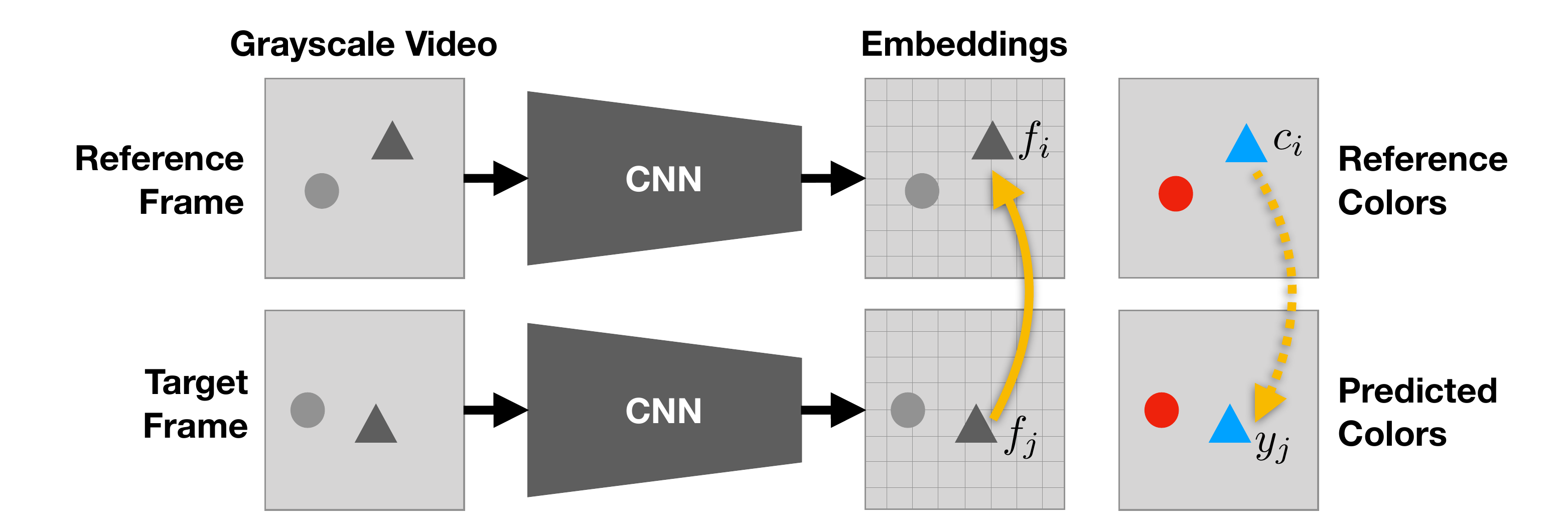}
    \caption{\textbf{Model Overview:} Given gray-scale frames, the model computes low-dimensional embeddings for each  location with a CNN. Using softmax similarity, the model points from the target frame into the reference frame embeddings (solid yellow arrow). The model then copies the color back into the predicted frame (dashed yellow arrow). After learning, we use the pointing mechanism as a visual tracker. Note that the model's pointer is soft, but for illustrations purposes we draw it as a single arrow.}
    \label{fig:model}
\end{figure}

We first describe how to train our model for video colorization, then discuss how to use it for tracking. See Figure \ref{fig:model} for a high level illustration of our model.

\subsection{Model}

\eat{
By forcing the model to predict the color of future frames by pointing into a reference frame in the past and copying its colors, the model needs to learn track regions in order to copy the right colors from the right place. We hypothesize that this video colorization task will cause a visual tracker to automatically emerge (and experiments suggest happens). 
}

Let $c_i \in \mathbb{R}^d$ be the true color for pixel $i$ in the reference frame, and let $c_j  \in \mathbb{R}^d$ be the true color for a pixel $j$ in the target frame. We denote $y_j  \in \mathbb{R}^d$ as the model's prediction for $c_j$. The model predicts $y_j$ as a linear combination of colors in the reference frame:
\begin{align}
y_j = \sum_i A_{ij}  c_i
\label{eqn:prop}
\end{align}
where $A$ is a similarity matrix between the target and reference frame such that the rows sum to one. Several similarity metrics are possible. We use inner product similarity normalized by softmax:
\begin{align}
A_{ij} = \frac{\exp\left(f_i^T f_j\right)}{\sum_k \exp\left(f_k^T f_j\right)}
\label{eqn:sim}
\end{align}
where $f_i \in \mathbb{R}^D$ is a low-dimensional embedding for pixel $i$ that is estimated by a convolutional neural network. Since we are computing distances between all pairs, the similarity matrix is potentially large. However, because color is fairly low spatial frequency, we can operate with lower resolution video frames allowing us to calculate and store all pairs on commodity hardware.

Similarity in color space does not imply that the embeddings are similar. Due to the softmax, the model only needs to point to one reference pixel in order to copy a color. Consequently, if there are two objects with the same color, the model does not constrain them to have the same embedding. This property enables the model to track multiple objects of the same color (which experiments show happens). 

Our model uses a pointing mechanism similar to attention networks \cite{bahdanau2014neural},  matching networks \cite{vinyals2016matching}, and pointer networks \cite{vinyals2015pointer}. However, our approach is unsupervised and we train the model for the purpose of using the underlying pointer mechanism as a visual tracker. Our model points within a single training example rather than across training examples.

\subsection{Learning}

\begin{figure}[tb]
    \centering
    \includegraphics[width=\linewidth]{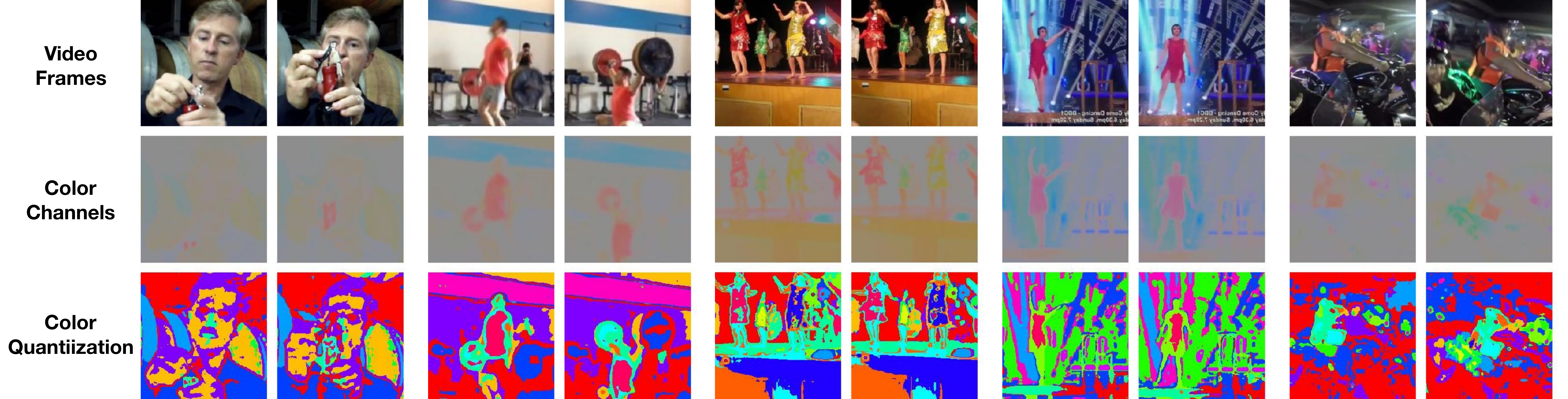}
    \caption{\textbf{Temporal Coherence of Color:} We visualize frames one second apart from the Kinetics training set \cite{kay2017kinetics}. The first row shows the original frames, and the second row shows the $ab$ color channels from $Lab$ space. The third row quantizes the color space into discrete bins and perturbs the colors to make the effect more pronounced. Unlabeled video from the web often has temporally coherent color, which provides excellent, large-scale training data for learning visual trackers. The last column shows an exception where a green light is turned on. Figure best viewed in color.}
    \label{fig:trainingdata}
\end{figure}

Our approach leverages the assumption during training that color is generally temporally stable. Clearly, there are exceptions, for example colorful lights can turn on and off. However, in practice, unlabeled video from the public web often has temporally stable color, which provides excellent, large-scale training data for learning to track. Figure \ref{fig:trainingdata} visualizes the coherency of color from a few videos on the Kinetics video dataset \cite{kay2017kinetics}.

We use a large dataset of unlabeled videos for learning. We train the parameters of the model $\theta$ such that the predicted colors $y_j$ are close to the target colors $c_j$ across the training set: 
\begin{align}
    \min_\theta \sum_j \mathcal{L}\left(y_j, c_j\right)
    \label{eqn:objective}
\end{align}
where $\mathcal{L}$ is the loss function. Since video colorization is a multi-modal problem \cite{zhang2016colorful}, we use the cross-entropy categorical loss after quantizing the color-space into discrete categories. We quantize by clustering the color channels across our dataset using $k$-means (we use $16$ clusters). We optimize Equation \ref{eqn:objective} using stochastic gradient descent.

\subsection{Inference}

After learning, we have a model that can compute
a similarity matrix $A$ for a pair of target and reference frames.
 Given an initially labeled frame from a held-out video, we use this pointer to propagate labels throughout the video. To do this, we exploit the property that our model is non-parametric in the label space. We simply re-use Equation \ref{eqn:prop} to propagate, but instead of propagating colors, we propagate distributions of categories. Since the rows of $A$ sum to one, Equation \ref{eqn:prop} can be interpreted as a mixture model where $A$ is the mixing coefficients. We will describe how to use this model for two different types of tasks: segment tracking and key-point tracking.

\textbf{Segment Tracking:} To track segments, we re-interpret $c_i \in \mathbb{R}^d$ as a vector indicating probabilities for $d$ categories. Note $d$ can change between learning/inference. In segmentation, the categories correspond to instances. We treat the background as just another category. The initial frame labels $c_i$ will be one-hot vectors (since we know the ground truth for the first frame), but the predictions $c_j$ in subsequent frames will be soft, indicating the confidence of the model. To make a hard decision, we can simply take the most confident category. 

\textbf{Keypoints Tracking:} Unlike colors and segmentation, keypoints are often sparse, but our model can still track them. We convert keypoints into a dense representation where $c_i \in \mathbb{R}^d$ is a binary vector indicating whether a keypoint is located at pixel $i$, if any. In this case, $d$ corresponds to the number of keypoints in the initial frame. We then proceed as we did in the segmentation case.

\begin{figure}[tb]
    \centering
    \includegraphics[width=0.9\linewidth]{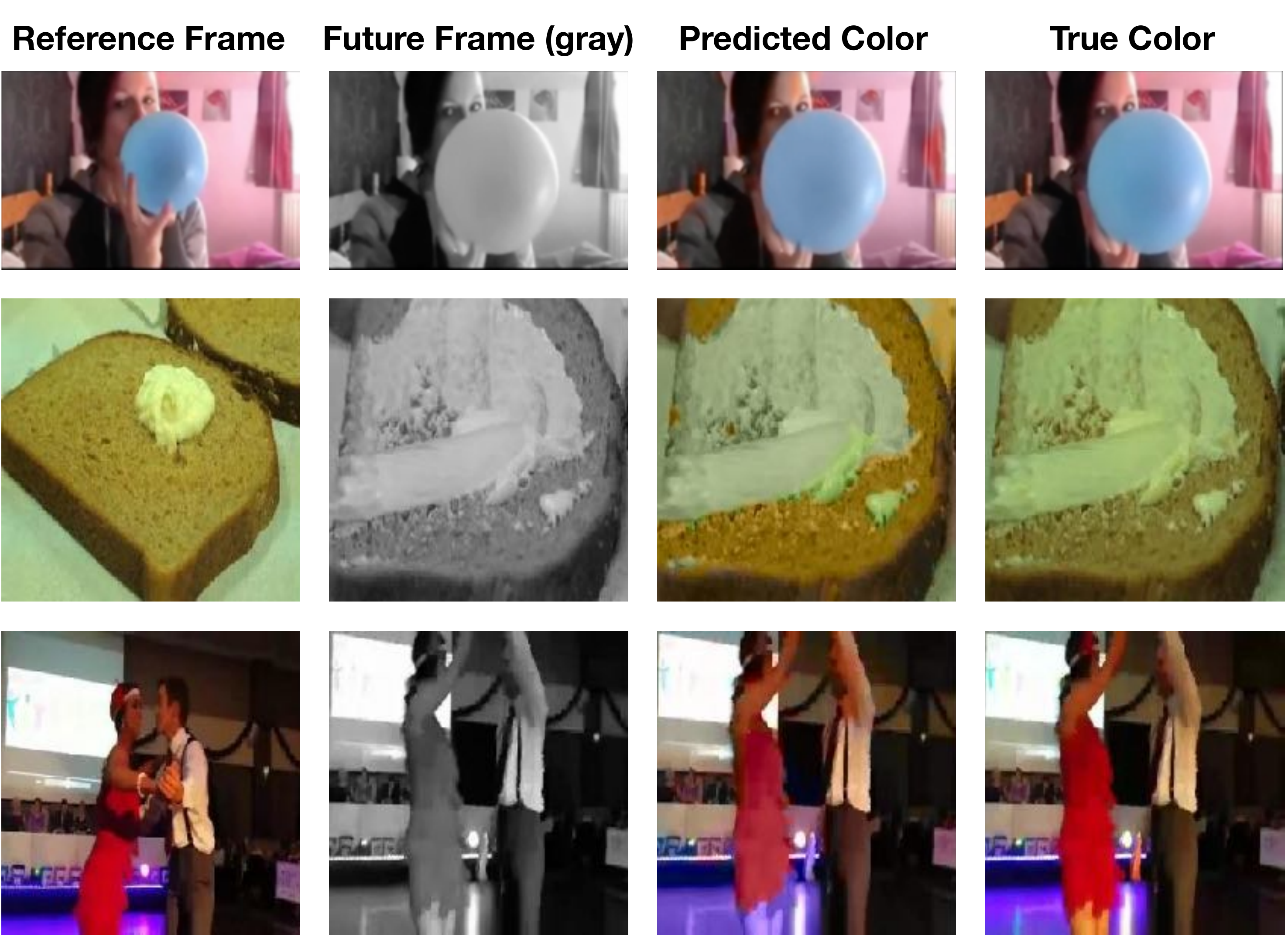}
    \caption{\textbf{Video Colorization:} We show video colorization results given a colorful reference frame. Our model learns to copy colors over many challenging transformations, such as butter spreading or people dancing. Best viewed in color.}
    \label{fig:videocolor}
\end{figure}

\textbf{Adjusting Temperature:} Equation \ref{eqn:prop} predicts a target label with a weighted average of all the labels in the reference frame. If the pointer is not confident, this can lead to blurry predictions over time, an effect also reported by \cite{zhang2016colorful}. To compensate for this, we can adjust the ``temperature'' of the softmax so that it makes more confident predictions. We simply divide the pre-softmax activations by a constant temperature $T$ during inference. Setting $T=1$ leaves the softmax distribution unchanged from training. We found $T=0.5$ works well for inference. 

\textbf{Variable Length Videos:}
During inference, we will be required to process long videos. We adopt a recursive approach in which we always propagate the labels given a window of previous $N$ frames (we use $N=3$).
Initially the window will contain the ground truth;
later it will contain the model's predictions.
\eat{
Given the previous $n$ video frames and their labels, we predict the labels for the next frame. We then feed these predictions back into the model to predict the subsequent labels.
}

\subsection{Implementation Details} 

We use a 3D convolutional network to produce $64$-dimensional embeddings. For efficiency, the network predicts a down-sampled feature map of $32 \times 32$ for each of the input frames. We use a ResNet-18 network architecture \cite{he2016deep} on each input frame, followed by a five layer 3D convolutional network. Note that to give the features global spatial information, we encode the spatial location as a two-dimensional vector in the range $[-1,1]$ and concatenate this to the features between the ResNet-18 and the 3D convolutional network.

The inputs to the model are four gray-scale video frames down-sampled to $256 \times 256$. We use the first three frames as reference frames, and the fourth frame as the target frame. The model pulls colors/labels from all three reference frames.
We pre-process the inputs to the network by scaling the intensities to be in the range $[-1,1]$, which is naturally near zero mean.
We use a frame rate of $6$ frames-per-second in learning and the full frame rate in inference.  
To quantize the color space, we convert the videos in the training set into $Lab$ space, take the $ab$ color channels, and cluster them with $k$-means. We represent the color of each pixel as a one-hot vector corresponding to the nearest cluster centroid.

We train our model for $400,000$ iterations. We use a batch size of $32$, and the Adam optimizer \cite{kingma2014adam}. We use a learning rate of $0.001$ for the first $60,000$ iterations and reduce it to $0.0001$ afterwards. The model is randomly initialized with Gaussian noise. Please see Appendix \ref{sec:networkarch} for more implementation details including network architecture. 

\section{Experiments}

The goal of our experiments to analyze how well a tracker can automatically emerge from our video colorization task. We first describe our experimental setup and baselines, then show two applications on video segmentation and human pose tracking. Finally, we visualize the embeddings learned by the model and analyze how to improve the tracker further. 

\begin{table}[tb]
    \small
    \centering
    \begin{tabular}{l | c | c  c }
        Method & Supervised?  & Segment  & Boundary  \\
        \hline
        Identity & & 22.1 & 23.6 \\
        Single Image Colorization &  & 4.7 & 5.2 \\
        Optical Flow (Coarse-to-Fine) \cite{liu2009beyond} & & 13.0 & 15.1  \\
        Optical Flow (FlowNet2) \cite{ilg2017flownet} & &  26.7 & 25.2 \\
        Ours & & 34.6 & 32.7  \\
        \hline
        Fully Supervised \cite{caelles2017one,yang2018efficient} & \checkmark  &  55.1 & 62.1  \\
    \end{tabular}
    \caption{\textbf{Video Segmentation Results}. We show performance on the DAVIS 2017 validation set for video segmentation.
        Higher numbers (which represent mean overlap) are better.
        We compare  against several baselines that do not use any labeled data during learning. Interestingly, our model learns a strong enough tracker to outperform optical flow based methods, suggesting that the model is learning useful motion and instance features.
        However, we still cannot yet match heavily supervised training.
        \eat{
    While more supervision generally leads to better performance, state-of-the-art results typically require training on large amounts of labeled data. In contrast, our model can learn to track even though it is trained without any ground truth labels.
}
    }
    \label{tab:davis2017}
\end{table}

\subsection{Experimental Setup}

We train our model on the training set from Kinetics \cite{kay2017kinetics}. Since our model learns from unlabeled video, we discard the labels. The Kinetics dataset is a large, diverse collection of $300,000$ videos from YouTube. We evaluate the model on the standard testing sets of other datasets depending on the task.  Since we are analyzing how well trackers emerge from video colorization, we compare against the following unsupervised baselines:

\textbf{Identity:} Since we are given labels for the initial testing frame, we have a baseline that assumes the video is static and repeats the initial label.

\textbf{Optical Flow:} We use state-of-the-art methods in optical flow as a baseline.  We experimented with two approaches. Firstly, we tried a classical optical flow implementation that is unsupervised and not learning based \cite{liu2009beyond}. Secondly, we also use a learning based approach that learns from synthetic data \cite{ilg2017flownet}. In both cases, we estimate  between frames and warp the initial labels to produce the predicted labels. We label a pixel as belonging to a category if the warped score is above a threshold. We experimented with several thresholds, and use the threshold that performs the best. We explored both recursive and non-recursive strategies, and report the strategy that works the best. Unless otherwise stated, we use the best performing optical flow based off FlowNet2 \cite{ilg2017flownet}.

\textbf{Single Image Colorization:} We evaluated how well computing similarity from the embeddings of a single image colorization model \cite{zhang2016colorful} work instead of our embeddings. Note this task is not designed nor originally intended for tracking by the authors. However, it allows us to quantify the difference between video and image colorization. To make this baseline, we train our model with the image colorization loss of \cite{zhang2016colorful}. We then follow the same tracking procedure, except using the features from the penultimate layer of the single image model for calculating similarity. 

\textbf{Supervised Models:} To analyze the gap between our self-supervised model and fully supervised approaches, we also consider the best available supervised approaches \cite{caelles2017one,yang2018efficient}.
Note that these methods train on
ImageNet, COCO segmentations, DAVIS, and even fine tune on the first frame of the test set.



\subsection{Video Colorization}

Figure \ref{fig:videocolor} shows example video colorization results given a reference frame, which is the task the model is originally trained on. We use the Kinetics validation set (not seen during training). The model learns to copy colors even across many challenging transformations, for example butter spreading on toast and people deforming as they dance. Since the model must copy colors from the reference frame, this suggests that the model may be robust to many difficult tracking situations. The rest of the section analyzes this tracking mechanism.

\subsection{Video Segmentation}
\label{sec:videoseg}

\begin{SCfigure}[2][tb]
    \centering
    \includegraphics[width=0.4\linewidth]{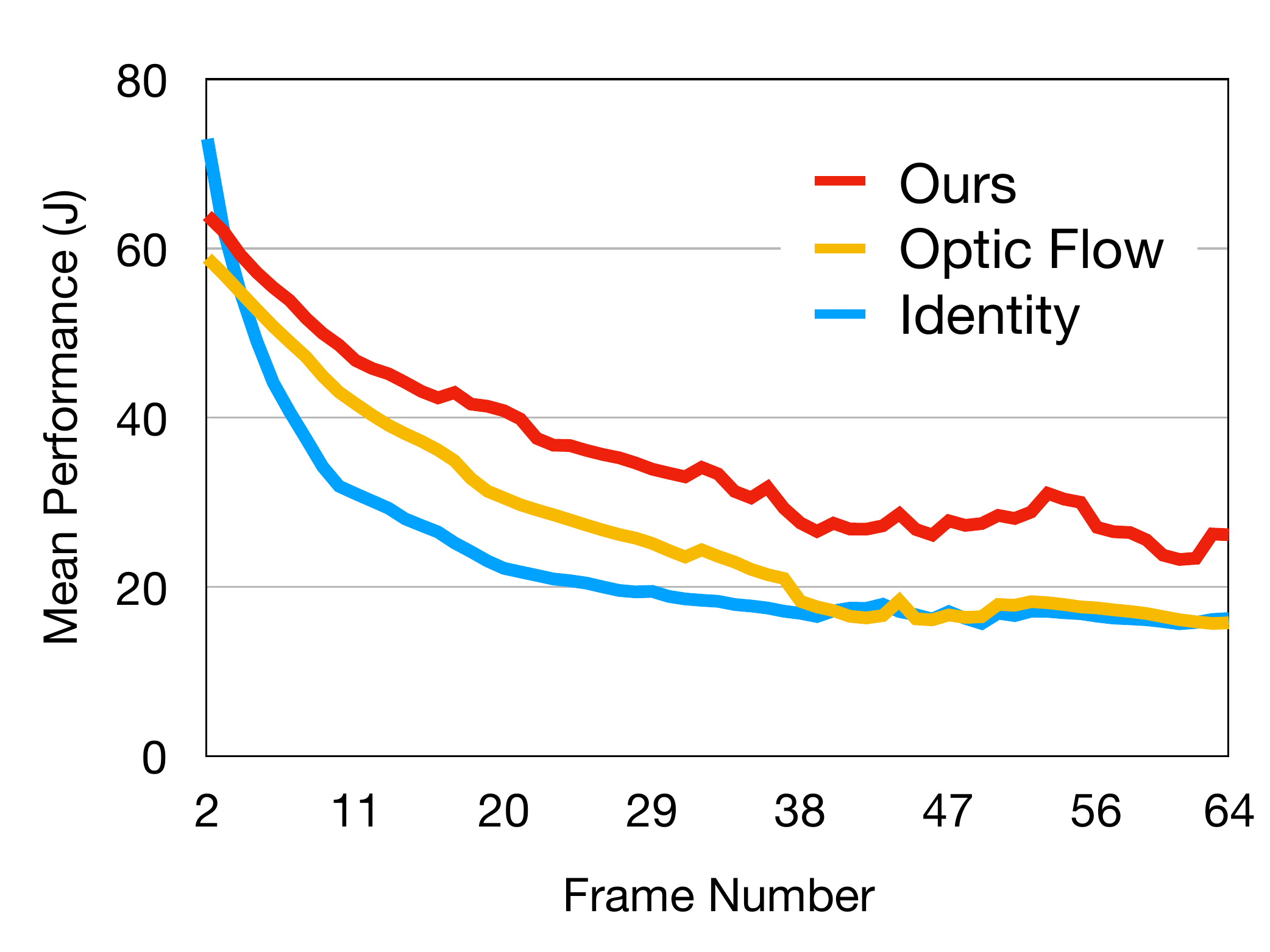}
    \caption{\textbf{Performance vs. Time:} We plot video segmentation average performance versus time in the video.  Our approach (red)  maintains more consistent performance for longer time periods than optical flow (orange). For long videos, optical flow on average degrades to the identity baseline. Since videos are variable length, we plot up to the median video length.}
    \label{fig:plotbytime}
\end{SCfigure}

\begin{figure}[tb]
    \centering
    \includegraphics[width=\linewidth]{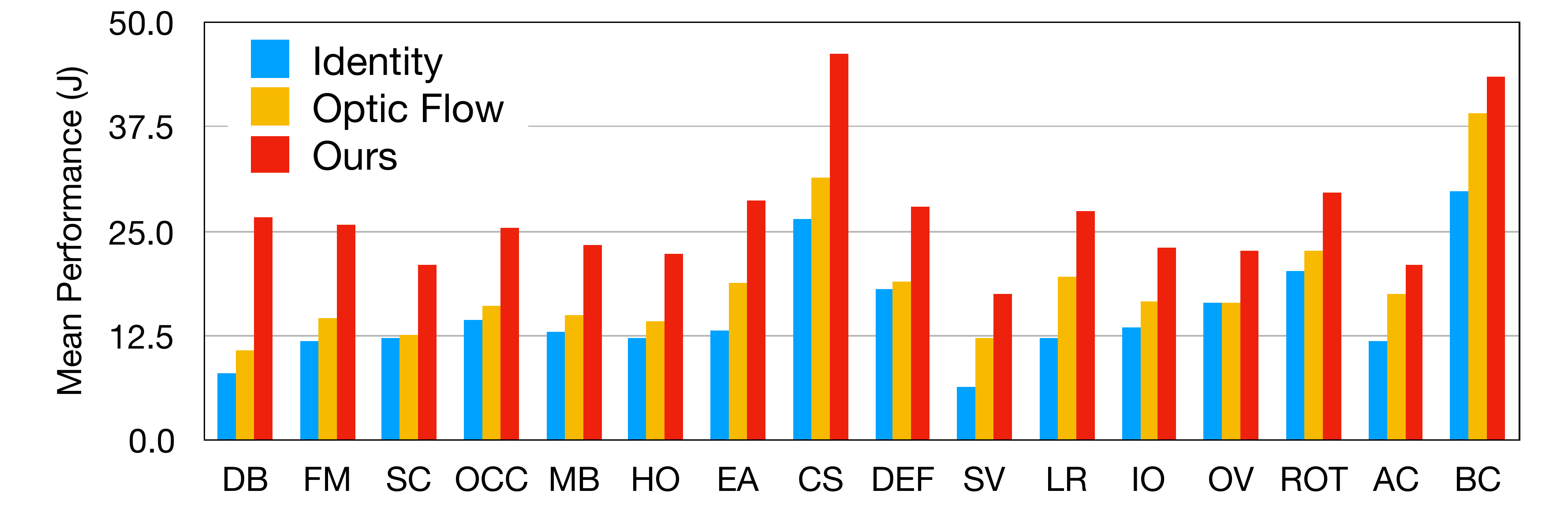}
    \caption{\textbf{Performance by Attribute:} We show the average performance broken down by attributes that describe the type of motion in the video. The attributes come from Table 1 of \cite{perazzi2016benchmark}. We sort the attributes by relative gain over optical flow. 
    }
    \label{fig:plotbyattr}
\end{figure}

We analyze our model on video segmentation with the
DAVIS 2017 validation set \cite{pont20182018} where the initial segmentation mask is given and the task is to predict the segmentation in the rest of the video. We follow the standard evaluation protocol using the validation set with the provided code and report two metrics that score segment overlap  and boundary accuracy. The videos in DAVIS 2017 are challenging and consist of multiple objects that undergo significant deformation, occlusion, and scale change with cluttered backgrounds.

\begin{figure}[p]
    \centering
    \includegraphics[width=\linewidth]{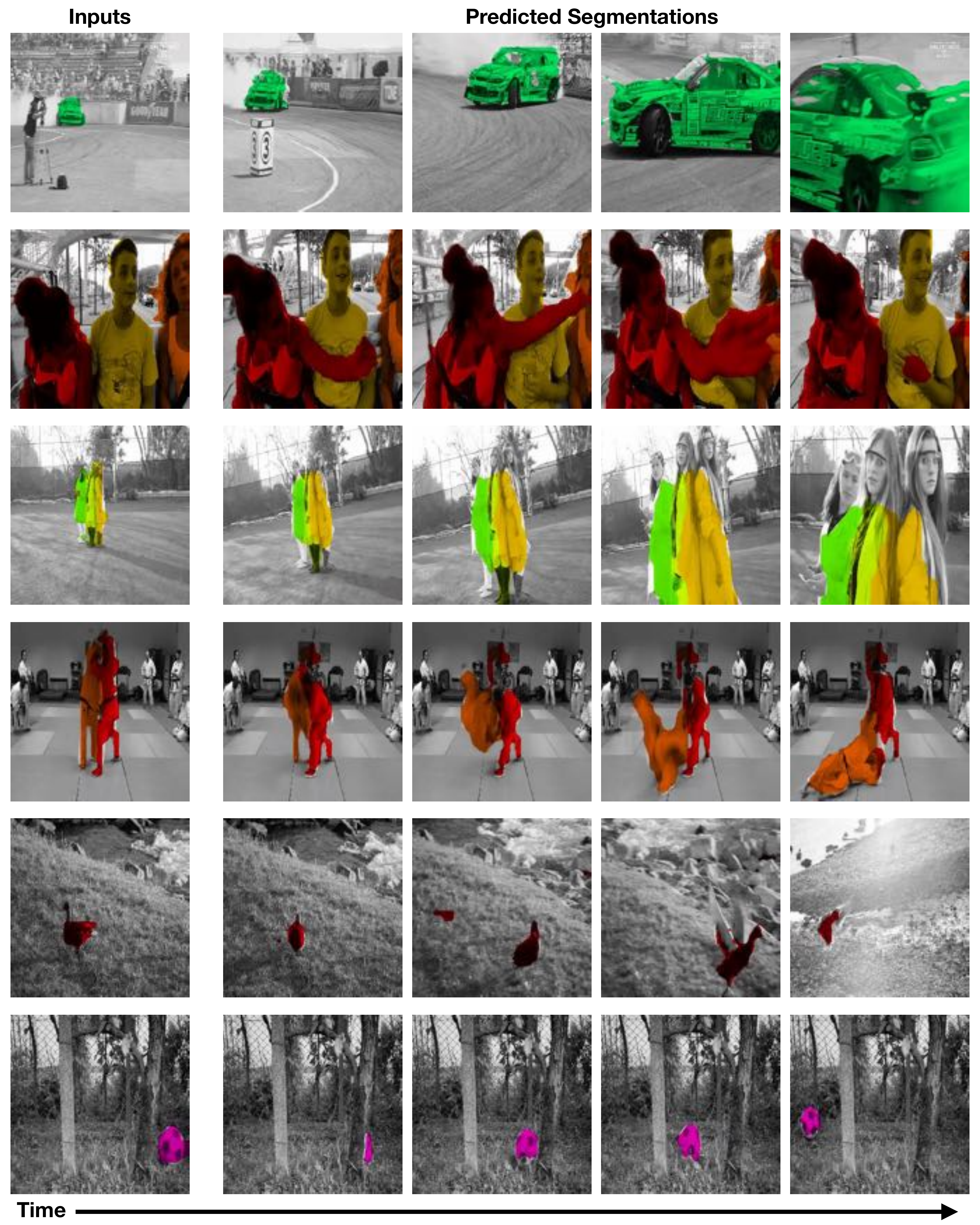}
    \caption{\textbf{Example Video Segmentations:} We show results from our self-supervised model on the task of video segmentation. Colors indicate different instances. Although the model is trained without ground truth labels, the model can still propagate segmentations throughout videos. The left column shows the input frame and input masks to the model, and the rest show the predictions. Results suggest that the model is generally robust to intra-class variations, such as deformations, and occlusions. The model often handles multiple objects and cluttered backgrounds. Best viewed in color.  We provide videos of results online at \textcolor{blue}{\url{https://goo.gl/qjHyPK}}}
    \label{fig:videosegresults}
\end{figure}

Table \ref{tab:davis2017} shows the performance on video segmentation. Our approach outperforms multiple methods in optical flow estimation. While the estimation of optical flow is often quite strong, warping the previous segment is challenging due to occlusion and motion blur. In contrast, our approach may excel because it also learns the warping mechanism end-to-end on video that contains an abundance of these challenging effects. 

We analyze how performance varies with the length of the video in Figure \ref{fig:plotbytime}. Our approach  maintains consistent performance for longer time periods than optical flow.  While optical flow works well in short time intervals, errors tend to accumulate over time. Our approach also has drift, but empirically colorization appears to learn more robust models. For long videos, optical flow based tracking eventually degrades to the identity baseline while ours remain relatively stronger for longer. The identity baseline, as expected, has a quick fall off as objects begin to move and deform.

We breakdown performance by video attributes in Figure \ref{fig:plotbyattr}. Our model tends to excel over optical flow for videos that have dynamic backgrounds (DB) and fast motion (FM), which are traditionally challenging situations for optical flow. Since our approach is trained end-to-end on videos that also have these artifacts, this suggests the model may be learning to handle the effects internally. Our model also shows strengths at cases involving occlusion (OCC) and motion blur (MB), which are difficult for optical flow because matching key-points is difficult under these conditions. Since color is low-frequency, it is not as affected by blur and occlusion during training. The most challenging situations for both our model and optical flow are due to scale variation (SV).

\begin{table}[tb!]
    \small
    \centering
    \begin{tabular}{l | c | c | c | c | c}
    Method & PCK@.1 & PCK@.2 & PCK@.3 & PCK@.4 & PCK@.5 \\
    \hline
    Identity & 43.1 & 64.5 & 76.0 & 83.5 & 88.5 \\
    Optical Flow (FlowNet2) \cite{ilg2017flownet} & 45.2 & 62.9 & 73.5 & 80.6 & 85.5 \\ 
    Ours & 45.2 & 69.6 & 80.8 & 87.5 & 91.4 \\
    \end{tabular}
    \caption{\textbf{Human Pose Tracking (no supervision):} We show performance on the JHMDB validation set for tracking human pose. PCK@X is the Probability of Correct Keypoint at a threshold of $X$ (higher numbers are better). At a strict threshold, our model tracks key-points with a similar performance as optical flow, suggesting that it is learning some motion features. At relaxed thresholds, our approach outperforms optical flow based methods, suggesting the errors caused by our model are less severe.
    }
    \label{tab:jhmdb}
\end{table}

\begin{figure}[tb!]
    \centering
    \includegraphics[width=0.8\linewidth]{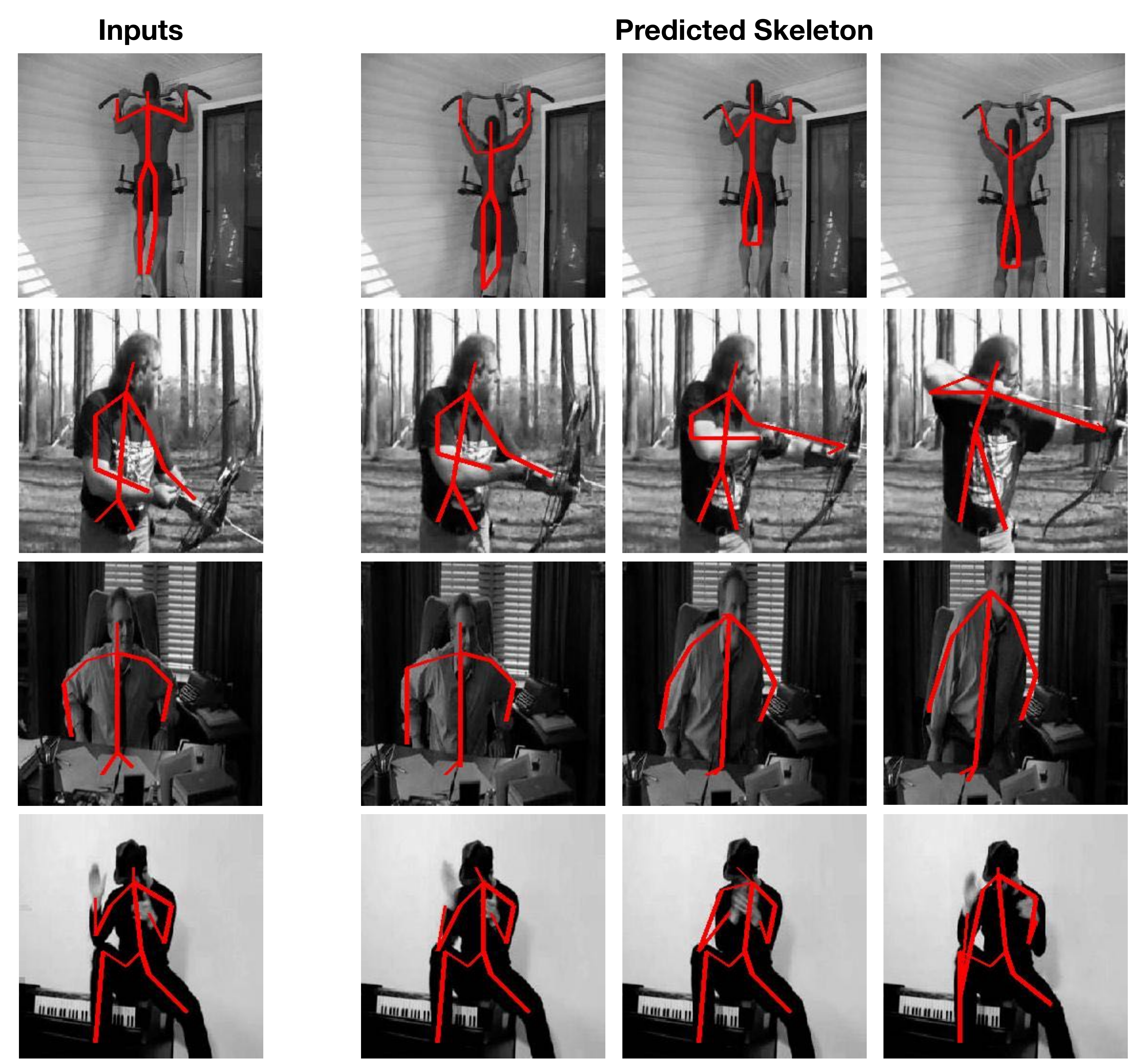}
    \caption{\textbf{Human Pose Tracking:} We show results from our self-supervised model for tracking human pose key-points. Although the model is trained without ground truth labels, the model can propagate skeletons labeled in the first frame throughout the rest of the video. Best viewed in color.}
    \label{fig:pose}
\end{figure}

To get a sense of the predicted segmentations,
Figure \ref{fig:videosegresults} shows a few example videos and the predicted segmentations from our method. Our model can successfully track multiple instances throughout the video, even when the objects are spatially near and have similar colors, for example the scene where multiple people are wearing similar white coats (third row). To quantify this, we analyze performance only on the videos with multiple objects (ranging from two to five objects). Under this condition, our model scores $31.0$ on segment overlap (J) versus $19.1$ for the optical flow based approach, suggesting our method still obtains strong performance with multiple objects. Finally, our model shows robustness to large deformations (second row) as well as large occlusions (second to last row). Typical failures include small objects and lack of fine-grained details.

\subsection{Pose Tracking}

\begin{figure}[t]
    \centering
    \includegraphics[width=0.8\linewidth]{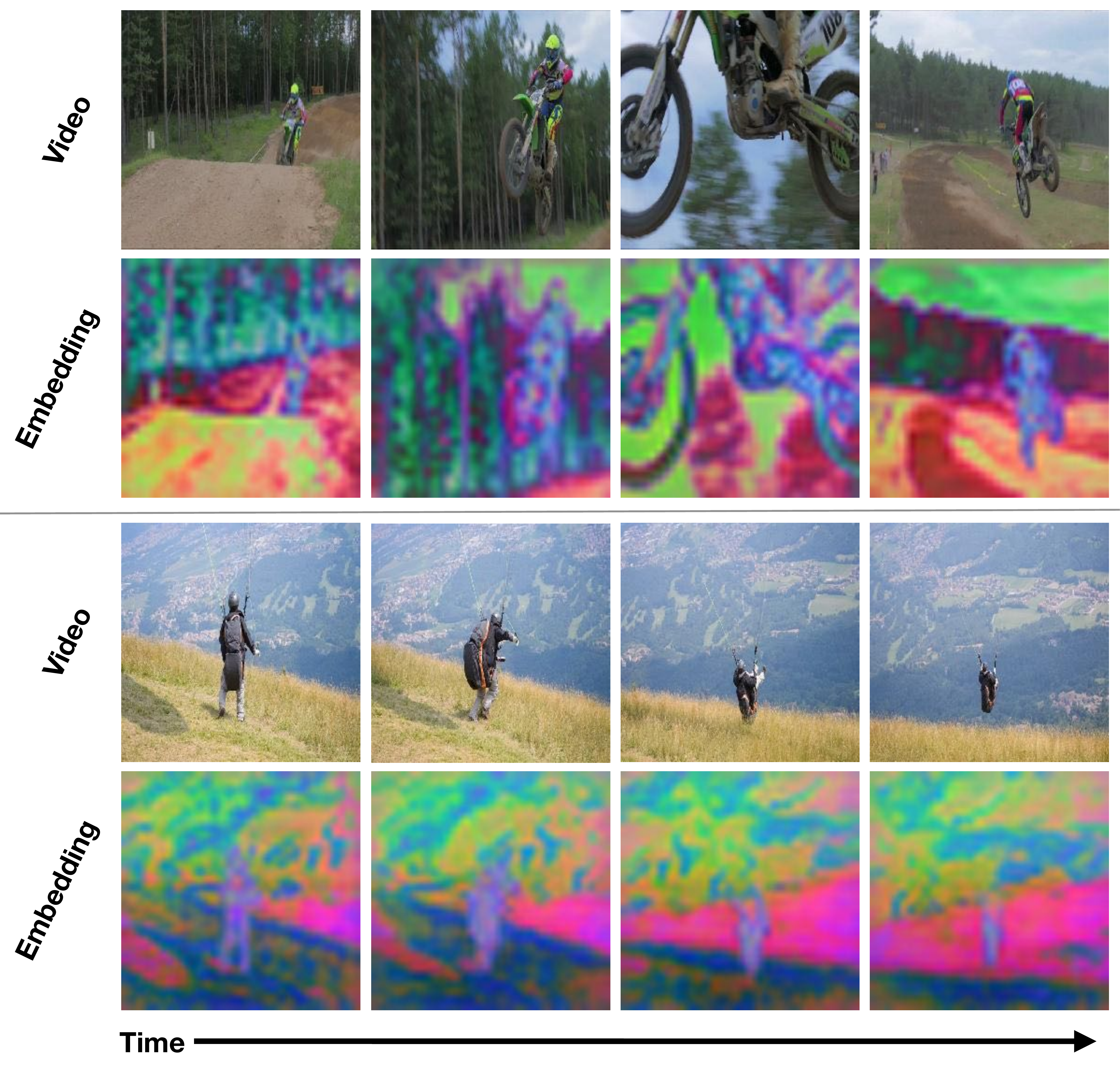}
    \caption{\textbf{Visualizing the Learned Embedding:} We project the embeddings into $3$ dimensions using PCA and visualize it as an RGB image. Similar colors illustrate the similarity in embedding space. Notice that the learned embeddings are stable over time even with significant deformation and viewpoint change. Best viewed in color.}
    \label{fig:embed_vis}
\end{figure}

\begin{figure}[t]
    \centering
    \includegraphics[width=0.8\linewidth]{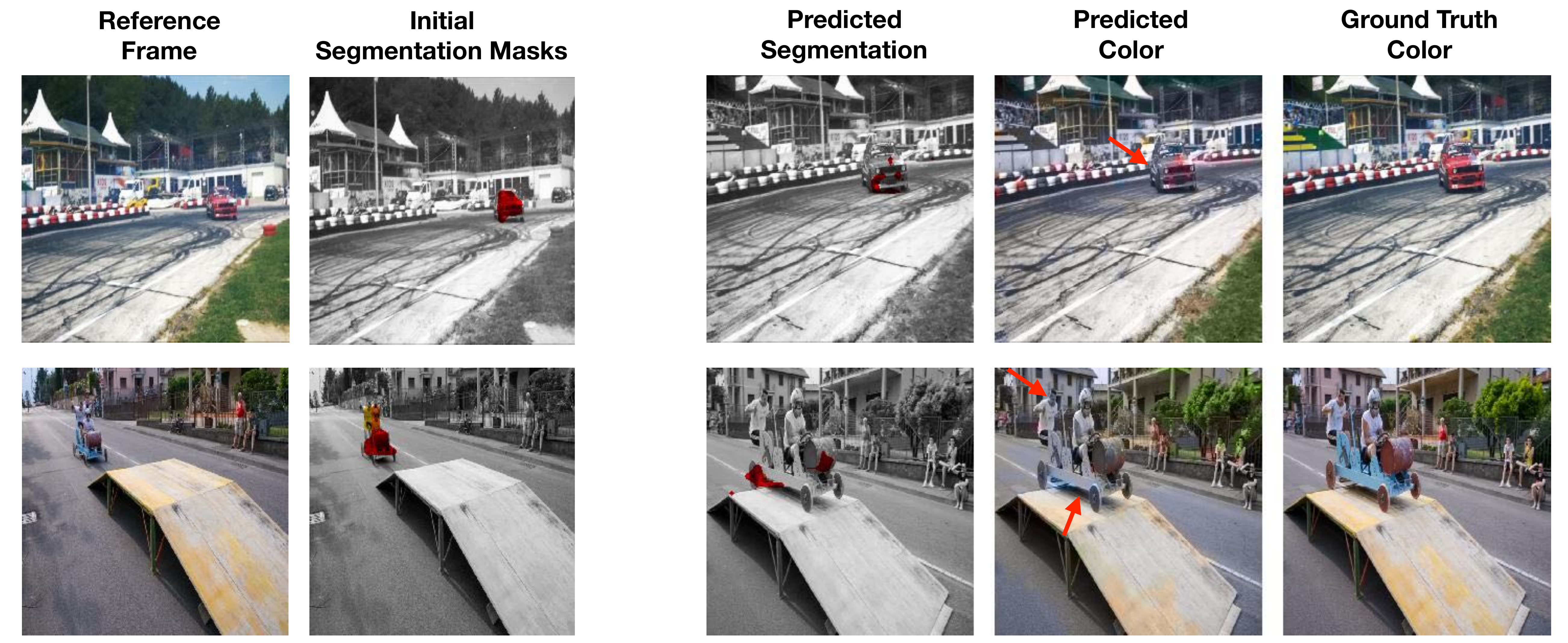}
    \caption{\textbf{Colorization vs.\ Tracking:} We show a few failures case where we do not correctly track the object, and the model also fails to propagate the colors (red arrows). This suggest that improving video colorization may translate into further improvements for self-supervised visual trackers. Best viewed in color.}
    \label{fig:failures}
\end{figure}

We experiment on human pose tracking with the JHMDB dataset \cite{Jhuang:ICCV:2013}. During testing, we are given an initial frame labeled with human keypoints and the task is to predict the keypoints in the subsequent frames. This task is challenging because it requires fine-grained  localization of keypoints when people undergo deformation. We use the standard PCK metric from \cite{yang2013articulated} which measures the percentage of keypoints that are sufficiently close to the ground truth. Following standard practice, we normalize the scale of the person. We normalize by the size of the person bounding box, and we report results at multiple threshold values X denoted as PCK@X. For more details, please see \cite{yang2013articulated}.

Table \ref{tab:jhmdb} shows the performance of our tracker versus baselines for tracking human pose given an initially labeled frame. At the most strict evaluation threshold, our model obtains similar performance to optical flow, suggesting that our model may be learning some motion features. At more relaxed thresholds, our model outperforms optical flow. This shows that the errors from optical flow tend to be more extreme than the errors from our tracker, even when the localization is not perfect. Moreover, the optical flow method is trained on large amounts of synthetic data, while our approach only requires video that is naturally available.

Figure \ref{fig:pose} shows qualitative results from our model on tracking human key-points. The model often can track large motions fairly well, such as the second and third row. Typical failures from the model are due to occlusion since a keypoint cannot be recovered once it disappears from the frame. 
\eat{
We expect that building models that colorize videos for longer time frames will help be robust to occlusion effects. Fortunately, the video colorization task provides a large-scale dataset ``for free.''
}

\subsection{Analysis of the model and its failure modes}

Since our model is trained on large amounts of unlabeled video, we are interested in gaining insight into what the model internally learns. Figure \ref{fig:embed_vis} visualizes the embeddings $f_i$ learned by our model by projecting them down to three dimensions using PCA and plotting it as an RGB image. The results show that nearest neighbors in the learned embedding space tend to correspond to object instances, even over significant deformations and viewpoint changes.

While our experiments show that these embeddings are useful for tracking,  there are still failures.
For example, Figure \ref{fig:failures} shows predicted segmentations from our tracker and the corresponding predicted colors. Moreover, we find that many of the failures to track are also failures to colorize.  To quantify this correlation, if any, we use the odds ratio between the two events of tracker failure and colorization failure. If the events are independent, we expect the odds ratio to be $1$. However, the odds ratio is $2.3$, suggesting moderate association. This suggests that there is still ``juice'' left in the video colorization signal for learning to track. We expect that building more accurate models for video colorization will translate into tracking improvements.

\section{Conclusion}

This paper shows that the task of video colorization is a promising signal for learning to track without requiring human supervision.
Our experiments show that learning to colorize video by pointing to a colorful reference frame causes a visual tracker to automatically emerge, which we leverage for video segmentation and human pose tracking. Moreover, our results suggest that improving the video colorization task may translate into improvements in self-supervised tracking. Since there is an abundance of unlabeled video in full color, video colorization appears to be a powerful signal for self-supervised learning of video models.

\appendix

\section{Performance by Attribute}

We reproduce Figure \ref{fig:plotbyattr} in the main paper, but expand the name of the attributes. This table shows the average performance broken down by attributes that describe the type of motion in the video. We sort the attributes by relative gain over optic flow. Please see Section \ref{sec:videoseg} for discussion.

\noindent
\includegraphics[width=\linewidth]{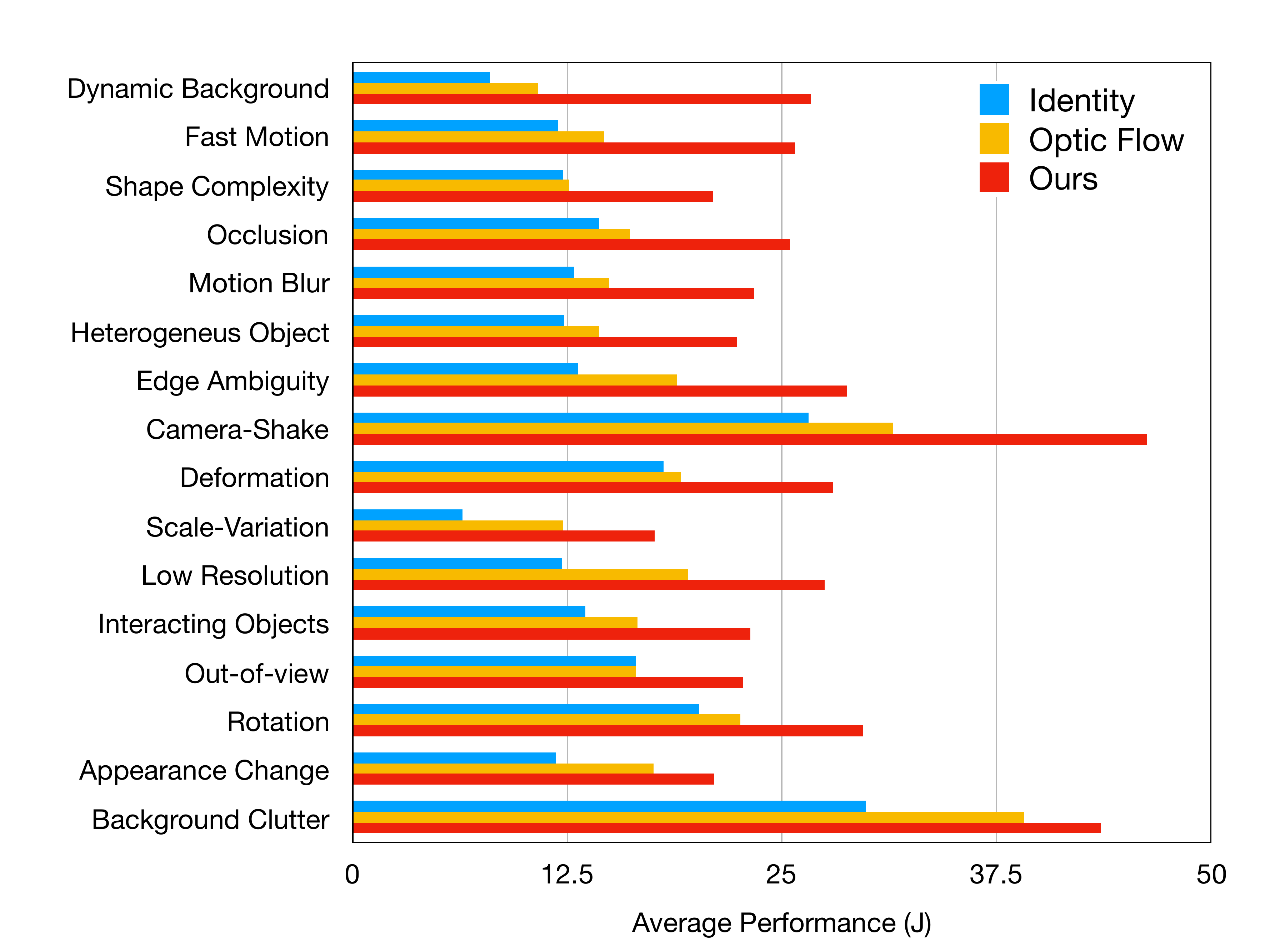}

\section{Network Architecture}
\label{sec:networkarch}

The inputs to the model are four gray-scale video frames down-sampled to $256 \times 256$.
We use a 3D convolutional network to produce $64$-dimensional embeddings. The network predicts a down-sampled feature map of $32 \times 32$ for each of the input frames.

We put a ResNet-18 network architecture on each input frame, following the original definition by \cite{he2016deep}. The ResNet-18 has weights that are shared across all frames. The only modification we make is to make the network output a  $32 \times 32$ spatial map of $256$ dimensions each. To do this, we remove the fully connected layers, global average pooling, and modify the output stride.

The features from the ResNet-18 backbone are then fed into a 3D spatio-temporal convolutional network. This network is defined as follows. We adopt Time $\times$ Width $\times$ Height notation.

\begin{center}
\begin{tabular}{l | c | c | c | c | c}
     Type & Kernel Size & Num Outputs & Stride & Padding & Dilation \\
    \hline
    Convolution & $1 \times 3 \times 3$ & 256 & 1 & 1 & $1 \times 1 \times 1$\\
    Convolution  & $3 \times 1 \times 1$ & 256 & 1 & 1 & $1 \times 1 \times 1$ \\
    Convolution  & $1 \times 3 \times 3$ & 256 & 1 & 1 & $1 \times 2 \times 2$\\
    Convolution  & $3 \times 1 \times 1$ & 256 & 1 & 1 & $1 \times 1 \times 1$ \\
    Convolution  & $1 \times 3 \times 3$ & 256 & 1 & 1 & $1 \times 4 \times 4$\\
    Convolution  & $3 \times 1 \times 1$ & 256 & 1 & 1 & $1 \times 1 \times 1$ \\
    Convolution  & $1 \times 3 \times 3$ & 256 & 1 & 1 & $1 \times 8 \times 8$\\
    Convolution  & $3 \times 1 \times 1$ & 256 & 1 & 1 & $1 \times 1 \times 1$ \\
    Convolution  & $1 \times 3 \times 3$ & 256 & 1 & 1 & $1 \times 16 \times 16$\\
    Convolution  & $3 \times 1 \times 1$ & 256 & 1 & 1 & $1 \times 1 \times 1$ \\
    Convolution  & $1 \times 1 \times 1$ & 64 & 1 & 1 & $1 \times 1 \times 1$ \\
\end{tabular}
\end{center}
\noindent Each convolution is followed by batch normalization and a rectiied linear unit (ReLU), except for the last layer, which produces the embeddings. Our implementation uses TensorFlow.

{
\bibliographystyle{splncs}
\bibliography{egbib}

\begin{thebibliography}{10}

\bibitem{kay2017kinetics}
Kay, W., Carreira, J., Simonyan, K., Zhang, B., Hillier, C., Vijayanarasimhan,
  S., Viola, F., Green, T., Back, T., Natsev, P., Suleyman, M., Zisserman, A.:
\newblock The kinetics human action video dataset.
\newblock arXiv preprint arXiv:1705.06950 (2017)

\bibitem{pont20182018}
Pont-Tuset, J., Caelles, S., Perazzi, F., Montes, A., Maninis, K.K., Chen, Y.,
  Van~Gool, L.:
\newblock The 2017 davis challenge on video object segmentation.
\newblock arXiv preprint arXiv:1803.00557 (2017)

\bibitem{Jhuang:ICCV:2013}
Jhuang, H., Gall, J., Zuffi, S., Schmid, C., Black, M.J.:
\newblock Towards understanding action recognition.
\newblock In: International Conf. on Computer Vision (ICCV). (December 2013)
  3192--3199

\bibitem{doersch2015unsupervised}
Doersch, C., Gupta, A., Efros, A.A.:
\newblock Unsupervised visual representation learning by context prediction.
\newblock In: Proceedings of the IEEE International Conference on Computer
  Vision. (2015)  1422--1430

\bibitem{owens2016ambient}
Owens, A., Wu, J., McDermott, J.H., Freeman, W.T., Torralba, A.:
\newblock Ambient sound provides supervision for visual learning.
\newblock In: European Conference on Computer Vision, Springer (2016)  801--816

\bibitem{jayaraman2015learning}
Jayaraman, D., Grauman, K.:
\newblock Learning image representations tied to ego-motion.
\newblock In: Proceedings of the IEEE International Conference on Computer
  Vision. (2015)  1413--1421

\bibitem{doersch2017multi}
Doersch, C., Zisserman, A.:
\newblock Multi-task self-supervised visual learning.
\newblock In: The IEEE International Conference on Computer Vision (ICCV).
  (2017)

\bibitem{wang2017transitive}
Wang, X., He, K., Gupta, A.:
\newblock Transitive invariance for self-supervised visual representation
  learning.
\newblock arXiv preprint arXiv:1708.02901 (2017)

\bibitem{zhang2017split}
Zhang, R., Isola, P., Efros, A.A.:
\newblock Split-brain autoencoders: Unsupervised learning by cross-channel
  prediction

\bibitem{larsson2017colorization}
Larsson, G., Maire, M., Shakhnarovich, G.:
\newblock Colorization as a proxy task for visual understanding.
\newblock In: CVPR. Volume~2. (2017) ~8

\bibitem{pathak2016context}
Pathak, D., Krahenbuhl, P., Donahue, J., Darrell, T., Efros, A.A.:
\newblock Context encoders: Feature learning by inpainting.
\newblock In: Proceedings of the IEEE Conference on Computer Vision and Pattern
  Recognition. (2016)  2536--2544

\bibitem{wang2015unsupervised}
Wang, X., Gupta, A.:
\newblock Unsupervised learning of visual representations using videos.
\newblock arXiv preprint arXiv:1505.00687 (2015)

\bibitem{vondrick2016generating}
Vondrick, C., Pirsiavash, H., Torralba, A.:
\newblock Generating videos with scene dynamics.
\newblock In: Advances In Neural Information Processing Systems. (2016)
  613--621

\bibitem{noroozi2016unsupervised}
Noroozi, M., Favaro, P.:
\newblock Unsupervised learning of visual representations by solving jigsaw
  puzzles.
\newblock In: European Conference on Computer Vision, Springer (2016)  69--84

\bibitem{pathak2017learning}
Pathak, D., Girshick, R., Doll{\'a}r, P., Darrell, T., Hariharan, B.:
\newblock Learning features by watching objects move.
\newblock In: Proc. CVPR. Volume~2. (2017)

\bibitem{isola2017image}
Isola, P., Zhu, J.Y., Zhou, T., Efros, A.A.:
\newblock Image-to-image translation with conditional adversarial networks.
\newblock arXiv preprint (2017)

\bibitem{pinto2016curious}
Pinto, L., Gandhi, D., Han, Y., Park, Y.L., Gupta, A.:
\newblock The curious robot: Learning visual representations via physical
  interactions.
\newblock In: European Conference on Computer Vision, Springer (2016)  3--18

\bibitem{agrawal2016learning}
Agrawal, P., Nair, A.V., Abbeel, P., Malik, J., Levine, S.:
\newblock Learning to poke by poking: Experiential learning of intuitive
  physics.
\newblock In: Advances in Neural Information Processing Systems. (2016)
  5074--5082

\bibitem{wu2016physics}
Wu, J., Lim, J.J., Zhang, H., Tenenbaum, J.B., Freeman, W.T.:
\newblock Physics 101: Learning physical object properties from unlabeled
  videos.
\newblock In: BMVC. Volume~2. (2016) ~7

\bibitem{tung2017self}
Tung, H.Y., Tung, H.W., Yumer, E., Fragkiadaki, K.:
\newblock Self-supervised learning of motion capture.
\newblock In: Advances in Neural Information Processing Systems. (2017)
  5242--5252

\bibitem{zhou2017unsupervised}
Zhou, T., Brown, M., Snavely, N., Lowe, D.G.:
\newblock Unsupervised learning of depth and ego-motion from video.
\newblock In: CVPR. Volume~2. (2017) ~7

\bibitem{zhou2016learning}
Zhou, T., Krahenbuhl, P., Aubry, M., Huang, Q., Efros, A.A.:
\newblock Learning dense correspondence via 3d-guided cycle consistency.
\newblock In: Proceedings of the IEEE Conference on Computer Vision and Pattern
  Recognition. (2016)  117--126

\bibitem{ilg2017flownet}
Ilg, E., Mayer, N., Saikia, T., Keuper, M., Dosovitskiy, A., Brox, T.:
\newblock Flownet 2.0: Evolution of optical flow estimation with deep networks.
\newblock In: IEEE Conference on Computer Vision and Pattern Recognition
  (CVPR). Volume~2. (2017)

\bibitem{zhou2016view}
Zhou, T., Tulsiani, S., Sun, W., Malik, J., Efros, A.A.:
\newblock View synthesis by appearance flow.
\newblock In: European conference on computer vision, Springer (2016)  286--301

\bibitem{welsh2002transferring}
Welsh, T., Ashikhmin, M., Mueller, K.:
\newblock Transferring color to greyscale images.
\newblock In: ACM Transactions on Graphics (TOG). Volume~21., ACM (2002)
  277--280

\bibitem{gupta2012image}
Gupta, R.K., Chia, A.Y.S., Rajan, D., Ng, E.S., Zhiyong, H.:
\newblock Image colorization using similar images.
\newblock In: Proceedings of the 20th ACM international conference on
  Multimedia, ACM (2012)  369--378

\bibitem{liu2008intrinsic}
Liu, X., Wan, L., Qu, Y., Wong, T.T., Lin, S., Leung, C.S., Heng, P.A.:
\newblock Intrinsic colorization.
\newblock In: ACM Transactions on Graphics (TOG). Volume~27., ACM (2008)  152

\bibitem{chia2011semantic}
Chia, A.Y.S., Zhuo, S., Gupta, R.K., Tai, Y.W., Cho, S.Y., Tan, P., Lin, S.:
\newblock Semantic colorization with internet images.
\newblock In: ACM Transactions on Graphics (TOG). Volume~30., ACM (2011)  156

\bibitem{deshpande2015learning}
Deshpande, A., Rock, J., Forsyth, D.:
\newblock Learning large-scale automatic image colorization.
\newblock In: Proceedings of the IEEE International Conference on Computer
  Vision. (2015)  567--575

\bibitem{zhang2016colorful}
Zhang, R., Isola, P., Efros, A.A.:
\newblock Colorful image colorization.
\newblock In: European Conference on Computer Vision, Springer (2016)  649--666

\bibitem{larsson2016learning}
Larsson, G., Maire, M., Shakhnarovich, G.:
\newblock Learning representations for automatic colorization.
\newblock In: European Conference on Computer Vision, Springer (2016)  577--593

\bibitem{guadarrama2017pixcolor}
Guadarrama, S., Dahl, R., Bieber, D., Norouzi, M., Shlens, J., Murphy, K.:
\newblock Pixcolor: Pixel recursive colorization.
\newblock arXiv preprint arXiv:1705.07208 (2017)

\bibitem{iizuka2016let}
Iizuka, S., Simo-Serra, E., Ishikawa, H.:
\newblock Let there be color!: joint end-to-end learning of global and local
  image priors for automatic image colorization with simultaneous
  classification.
\newblock ACM Transactions on Graphics (TOG) \textbf{35}(4) (2016)  110

\bibitem{ironi2005colorization}
Ironi, R., Cohen-Or, D., Lischinski, D.:
\newblock Colorization by example.
\newblock In: Rendering Techniques, Citeseer (2005)  201--210

\bibitem{yatziv2006fast}
Yatziv, L., Sapiro, G.:
\newblock Fast image and video colorization using chrominance blending.
\newblock IEEE transactions on image processing \textbf{15}(5) (2006)
  1120--1129

\bibitem{heu2009image}
Heu, J.H., Hyun, D.Y., Kim, C.S., Lee, S.U.:
\newblock Image and video colorization based on prioritized source propagation.
\newblock In: Image Processing (ICIP), 2009 16th IEEE International Conference
  on, IEEE (2009)  465--468

\bibitem{liu2018switchable}
Liu, S., Zhong, G., De~Mello, S., Gu, J., Yang, M.H., Kautz, J.:
\newblock Switchable temporal propagation network.
\newblock arXiv preprint arXiv:1804.08758 (2018)

\bibitem{badrinarayanan2010label}
Badrinarayanan, V., Galasso, F., Cipolla, R.:
\newblock Label propagation in video sequences.
\newblock In: Computer Vision and Pattern Recognition (CVPR), 2010 IEEE
  Conference on, IEEE (2010)  3265--3272

\bibitem{ramakanth2014seamseg}
Ramakanth, S.A., Babu, R.V.:
\newblock Seamseg: Video object segmentation using patch seams.
\newblock In: CVPR. Volume~2. (2014) ~5

\bibitem{vijayanarasimhan2012active}
Vijayanarasimhan, S., Grauman, K.:
\newblock Active frame selection for label propagation in videos.
\newblock In: European conference on computer vision, Springer (2012)  496--509

\bibitem{perazzi2015fully}
Perazzi, F., Wang, O., Gross, M., Sorkine-Hornung, A.:
\newblock Fully connected object proposals for video segmentation.
\newblock In: Proceedings of the IEEE international conference on computer
  vision. (2015)  3227--3234

\bibitem{grundmann2010efficient}
Grundmann, M., Kwatra, V., Han, M., Essa, I.:
\newblock Efficient hierarchical graph-based video segmentation.
\newblock In: Computer Vision and Pattern Recognition (CVPR), 2010 IEEE
  Conference on, IEEE (2010)  2141--2148

\bibitem{xu2012evaluation}
Xu, C., Corso, J.J.:
\newblock Evaluation of super-voxel methods for early video processing.
\newblock In: Computer Vision and Pattern Recognition (CVPR), 2012 IEEE
  Conference on, IEEE (2012)  1202--1209

\bibitem{brox2010object}
Brox, T., Malik, J.:
\newblock Object segmentation by long term analysis of point trajectories.
\newblock In: European conference on computer vision, Springer (2010)  282--295

\bibitem{fragkiadaki2012video}
Fragkiadaki, K., Zhang, G., Shi, J.:
\newblock Video segmentation by tracing discontinuities in a trajectory
  embedding.
\newblock In: Computer Vision and Pattern Recognition (CVPR), 2012 IEEE
  Conference on, IEEE (2012)  1846--1853

\bibitem{yang2018efficient}
Yang, L., Wang, Y., Xiong, X., Yang, J., Katsaggelos, A.K.:
\newblock Efficient video object segmentation via network modulation.
\newblock arXiv preprint arXiv:1802.01218 (2018)

\bibitem{caelles2017one}
Caelles, S., Maninis, K.K., Pont-Tuset, J., Leal-Taix{\'e}, L., Cremers, D.,
  Van~Gool, L.:
\newblock One-shot video object segmentation.
\newblock In: CVPR 2017, IEEE (2017)

\bibitem{perazzi2017learning}
Perazzi, F., Khoreva, A., Benenson, R., Schiele, B., Sorkine-Hornung, A.:
\newblock Learning video object segmentation from static images.
\newblock In: Computer Vision and Pattern Recognition. (2017)

\bibitem{deng2009imagenet}
Deng, J., Dong, W., Socher, R., Li, L.J., Li, K., Fei-Fei, L.:
\newblock Imagenet: A large-scale hierarchical image database.
\newblock In: Computer Vision and Pattern Recognition, 2009. CVPR 2009. IEEE
  Conference on, IEEE (2009)  248--255

\bibitem{lin2014microsoft}
Lin, T.Y., Maire, M., Belongie, S., Hays, J., Perona, P., Ramanan, D.,
  Doll{\'a}r, P., Zitnick, C.L.:
\newblock Microsoft coco: Common objects in context.
\newblock In: European conference on computer vision, Springer (2014)  740--755

\bibitem{faktor2014video}
Faktor, A., Irani, M.:
\newblock Video segmentation by non-local consensus voting.
\newblock In: BMVC. Volume~2. (2014) ~8

\bibitem{marki2016bilateral}
M{\"a}rki, N., Perazzi, F., Wang, O., Sorkine-Hornung, A.:
\newblock Bilateral space video segmentation.
\newblock In: Proceedings of the IEEE Conference on Computer Vision and Pattern
  Recognition. (2016)  743--751

\bibitem{khoreva2017lucid}
Khoreva, A., Benenson, R., Ilg, E., Brox, T., Schiele, B.:
\newblock Lucid data dreaming for multiple object tracking.
\newblock arXiv preprint arXiv:1703.09554 (2017)

\bibitem{bahdanau2014neural}
Bahdanau, D., Cho, K., Bengio, Y.:
\newblock Neural machine translation by jointly learning to align and
  translate.
\newblock arXiv preprint arXiv:1409.0473 (2014)

\bibitem{vinyals2016matching}
Vinyals, O., Blundell, C., Lillicrap, T., Wierstra, D.,  et~al.:
\newblock Matching networks for one shot learning.
\newblock In: Advances in Neural Information Processing Systems. (2016)
  3630--3638

\bibitem{vinyals2015pointer}
Vinyals, O., Fortunato, M., Jaitly, N.:
\newblock Pointer networks.
\newblock In: Advances in Neural Information Processing Systems. (2015)
  2692--2700

\bibitem{he2016deep}
He, K., Zhang, X., Ren, S., Sun, J.:
\newblock Deep residual learning for image recognition.
\newblock In: Proceedings of the IEEE conference on computer vision and pattern
  recognition. (2016)  770--778

\bibitem{kingma2014adam}
Kingma, D.P., Ba, J.:
\newblock Adam: A method for stochastic optimization.
\newblock arXiv preprint arXiv:1412.6980 (2014)

\bibitem{liu2009beyond}
Liu, C.,  et~al.:
\newblock Beyond pixels: exploring new representations and applications for
  motion analysis.
\newblock PhD thesis, Massachusetts Institute of Technology (2009)

\bibitem{perazzi2016benchmark}
Perazzi, F., Pont-Tuset, J., McWilliams, B., Van~Gool, L., Gross, M.,
  Sorkine-Hornung, A.:
\newblock A benchmark dataset and evaluation methodology for video object
  segmentation.
\newblock In: Proceedings of the IEEE Conference on Computer Vision and Pattern
  Recognition. (2016)  724--732

\bibitem{yang2013articulated}
Yang, Y., Ramanan, D.:
\newblock Articulated human detection with flexible mixtures of parts.
\newblock IEEE transactions on pattern analysis and machine intelligence
  \textbf{35}(12) (2013)  2878--2890

\end{thebibliography}
}

\end{document}